\documentclass{article} % For LaTeX2e
\usepackage[preprint]{colm2025_conference}
\usepackage{microtype}
\usepackage{hyperref}
\usepackage{url}
\usepackage{booktabs}
\usepackage{amsmath}    % For advanced math structures
\usepackage{amsfonts}   % For math fonts and \mathbb
\usepackage{amssymb}    % For additional math symbols
\usepackage{algorithm}
\usepackage{algpseudocode} % For the algorithmic environment
\usepackage{graphicx}  
\usepackage{lineno}
\usepackage{multirow}
\usepackage{xcolor}
\usepackage{wrapfig}
\usepackage{subcaption}
\usepackage{listings}
\usepackage[frozencache,cachedir=.]{minted}

\definecolor{darkblue}{rgb}{0, 0, 0.5}
\hypersetup{colorlinks=true, citecolor=darkblue, linkcolor=darkblue, urlcolor=darkblue}

\newcommand{\question}{q}
\newcommand{\answer}{a}
\newcommand{\rationale}{r}

\newcommand{\hlg}[1]{\textcolor{green!60!black}{#1}}
\newcommand{\hld}[1]{\textcolor{red!60!black}{#1}}

\title{\undofull}

\author{Kushal Jain\\
UC San Diego\\
\texttt{knjain@ucsd.edu} \\
\And
Piyushi Goyal \& Kumar Shridhar \\
ETH Zurich \\
\texttt{\{pgoyal,shkumar\}@ethz.ch} \\
}

\newcommand{\undofull}{\textsf{UNDO}:  \textbf{UN}derstanding \textbf{D}istillation as \textbf{O}ptimization}

\newcommand{\undo}{\textsf{UNDO}}

\begin{document}

\ifcolmsubmission
\linenumbers
\fi

\maketitle

\begin{abstract}
Knowledge distillation has emerged as an effective strategy for compressing large language models' (LLMs) knowledge into smaller, more efficient student models. However, standard one-shot distillation methods often produce suboptimal results due to a mismatch between teacher-generated rationales and the student's specific learning requirements. In this paper, we introduce the \undofull\ framework, designed to bridge this gap by iteratively identifying the student's errors and prompting the teacher to refine its explanations accordingly. Each iteration directly targets the student's learning deficiencies, motivating the teacher to provide tailored and enhanced rationales that specifically address these weaknesses. Empirical evaluations on various challenging mathematical and commonsense reasoning tasks demonstrate that our iterative distillation method, \undo, significantly outperforms standard one-step distillation methods, achieving performance gains of up to 20\%. Additionally, we show that teacher-generated data refined through our iterative process remains effective even when applied to different student models, underscoring the broad applicability of our approach. Our work fundamentally reframes knowledge distillation as an iterative \emph{teacher–student} interaction, effectively leveraging dynamic refinement by the teacher for better knowledge distillation.

\end{abstract} 

\section{Introduction}

Knowledge distillation has emerged as an effective strategy for compressing large language models (LLMs) into smaller, more efficient student models \citep{hinton2015distillingknowledgeneuralnetwork,sanh2020distilbertdistilledversionbert, 10.1016/j.neunet.2024.106594}. Initially, the primary approach involved aligning the student's outputs with the soft logits of the teacher, enabling the student to approximate the teacher’s distribution. Recently, this paradigm has expanded to include \emph{chain-of-thought} (CoT) reasoning, which seeks to instill smaller models with interpretable intermediate steps rather than simply matching final outputs \citep{shridhar-etal-2023-distilling, hsieh-etal-2023-distilling, li-etal-2023-symbolic, chenglin-etal-2024-mixed}.

Despite these advancements, standard one-shot distillation methods frequently suffer from a distributional mismatch: the rationales generated by the teacher often do not align well with the student's specific learning challenges \citep{agarwal2024onpolicy, adarsh2024siked}. To overcome this limitation, we introduce the \undofull\ framework, an \emph{iterative} distillation strategy that actively identifies and addresses the student's errors by continuously prompting the teacher to generate refined, targeted rationales. In each iteration, the teacher's generation and student’s responses are analyzed by the teacher, leading to tailored explanations that directly address the student's learning gaps. This loop ensures that the training data becomes more personalized to the student’s specific needs, instead of simply reflecting generic teacher behavior.

Our iterative approach is inspired by educational concepts such as \emph{scaffolding} \citep{wood1976role} and \emph{formative assessment} \citep{black2004working}, which highlight the importance of adaptive, targeted feedback in effective learning. Similar to a skilled teacher who adjusts instruction based on a learner’s struggles, our method dynamically identifies and addresses the student's errors, thereby maintaining the learner within their \emph{zone of proximal development} (ZPD) \citep{cole1978mind}.

Empirical evaluations conducted on challenging mathematical reasoning datasets, including MATH \citep{lightman2024lets} and MMLU Pro \citep{wang2024mmlupro}, demonstrate that our iterative distillation method significantly outperforms traditional one-shot distillation, delivering improvements of up to 20\%. Furthermore, we show that iteratively refined teacher-generated data maintains its effectiveness across diverse student models, highlighting the robustness and generality of our method. Finally, we show that student models trained using our proposed approach outperform standard distilled models on out-of-domain tasks. Our work reframes knowledge distillation as a dynamic, iterative \emph{teacher–student} interaction, leading to a better learning outcome for the student.

\section{Related Work}

\paragraph{Knowledge distillation}
Extensive research has explored knowledge distillation since \citet{hinton2015distillingknowledgeneuralnetwork} introduced the concept of training smaller student models using soft-label outputs (soft logits) from larger teacher models. Early influential models such as DistilBERT \citep{sanh2020distilbertdistilledversionbert} demonstrated substantial parameter reductions (~40\%) without significant performance losses. Subsequent advancements, like MiniLM \citep{NEURIPS2020_3f5ee243}, further refined this technique by focusing specifically on distilling the self-attention mechanism. More recent works, including \citet{shum-etal-2024-first} and \citet{timiryasov-tastet-2023-baby}, explored selective training strategies by choosing essential samples from the teacher’s training set.

The integration of Chain-of-Thought (CoT) reasoning \citep{wei2022chain} into distillation has sparked additional innovations. For example, \citet{shridhar-etal-2023-distilling} trained student models to explicitly replicate the teacher's CoT reasoning steps. Similarly, \citet{li-etal-2023-symbolic} generated diverse rationales for individual questions, while \citet{chenglin-etal-2024-mixed} extended this by incorporating multiple reasoning styles. \citet{10.1016/j.neunet.2024.106594} distilled mathematical reasoning into structured equations and combined various reasoning modalities (text, code, equations). Further, \citet{luo2025wizardmath} leveraged reinforcement learning to fine-tune student models. Despite these advancements, most approaches remain one-directional, with teacher outputs independent of student-specific needs. In contrast, our method explicitly addresses the student’s learning gaps by encouraging teacher awareness and adaptation.

Surveys such as \citet{Gou_2021} and \citet{10.1145/3699518} have also comprehensively discussed the benefits and diverse applications of knowledge distillation across multiple domains, including computer vision and natural language processing.

\paragraph{Iterative distillation}
Our work draws significant inspiration from iterative forms of knowledge distillation. Previous research often emphasized the transfer of intermediate reasoning steps to smaller models, either sequentially \citep{hsieh-etal-2023-distilling,magister-etal-2023-teaching, shridhar-etal-2023-distilling} or concurrently integrating various reasoning strategies \citep{chenglin-etal-2024-mixed, 10.1016/j.neunet.2024.106594, jain2024firststepadvantageimportancestarting}. Notably, \citet{wang2023democratizing} employed iterative interactions between students and black-box teachers, with teachers providing targeted feedback based on student failures. Similarly, \citet{adarsh2024siked} combined iterative self-guided training and self-generated data to enhance reasoning in smaller models. Additionally, \citet{agarwal2024onpolicy} utilized on-policy training to better align student and teacher distributions by incorporating student-generated outputs. Self-distillation methods \citep{zelikman2022star, liu2024smart} involved models generating, filtering, and fine-tuning based on their own rationales. Reinforcement learning-based self-training was further explored by \citet{gulcehre2023reinforcedselftrainingrestlanguage}, where off-policy sampling informed dataset generation. Unlike these methods, our iterative approach uniquely emphasizes a dynamic, adaptive feedback loop where the teacher explicitly targets and rectifies the student's learning challenges.

\begin{figure}
    \centering
    \includegraphics[width=0.99\linewidth]{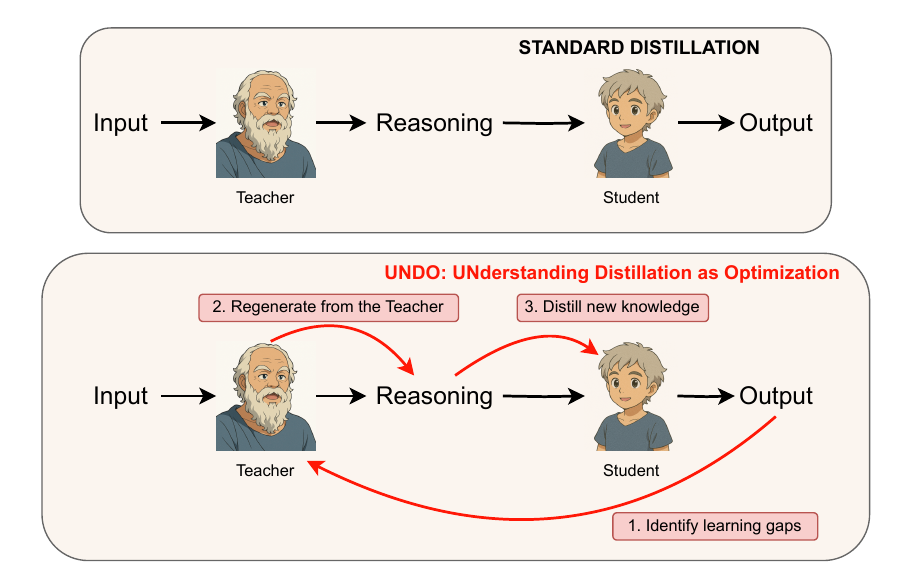}
    \vspace{1\baselineskip}
    \caption{We compare \undofull (bottom) with Standard Distillation (top). The main difference lies in the iterative nature of \undo where once a knowledge is distilled into the student, the teacher iteratively \emph{identifies the learning gap}, \emph{regenerate its reasoning chains} and \emph{distill the knowledge} again.}
    \label{fig:main-fig}
\end{figure}
\vspace{1\baselineskip}

\section{Methodology}

In this section, we first outline the standard knowledge distillation process and discuss its key limitations. We then introduce our proposed \undofull\ framework, which addresses these limitations by enabling the teacher to iteratively identify the student's weaknesses and refine the knowledge being distilled accordingly. 

\paragraph{Problem Setup.}
We consider a large teacher model $p_L$ (LLM) and a smaller student model $p_{\texttt{sm}}^\theta$ (with learnable parameters~$\theta$). We have a dataset of math questions $\mathcal{D} = \{(\question_i, \answer_i)\}_{i=1}^{n}$, and an associated held-out validation set $\mathcal{V} = \{(\question_j, \answer_j)\}_{j=1}^{m}$ to monitor progress at each iteration. Our goal is to distill the teacher's reasoning capabilities into the student iteratively by constantly moderating the errors made by the student and targeted refinements from the teacher.

\subsection{Standard Knowledge Distillation}
We begin with the standard knowledge distillation approach following \citet{shridhar-etal-2023-distilling, magister-etal-2023-teaching}, where the teacher model $p_L$ is prompted to produce a rationale (chain-of-thought) $\rationale_i$ for each question $\question_i$. Let $\hat{\answer}_i$ be the final numeric result in $\rationale_i$. We only keep those samples for which $\hat{\answer}_i$ matches the ground truth $\answer_i$. This produces a filtered dataset $\mathcal{D}_{\texttt{LLM}}=\{(\question_i,\rationale_i)\}_{i=1}^{N}$ of correct (question, rationale) pairs. 

The student is then fine-tuned by maximizing the log-likelihood of these rationales:
\begin{equation*}
    \label{eq:one_step_nll}
    \begin{split}
        \mathcal{L}_{\texttt{L}}(\theta) 
        \;=\;& - \mathbb{E}_{(\question_i,\, \rationale_i)\,\sim\,\mathcal{D}_{\texttt{LLM}}} \Bigg[ 
        \sum_{t=1}^{M_i} \log\,p_{\texttt{sm}}^\theta \bigl(\rationale_{i,t} \,\mid\, \rationale_{i,<t},\,\question_i,\, I \bigr)\Bigg],
    \end{split}
\end{equation*}
where $M_i$ is the number of tokens of the rationale, and $I$ is an instruction used to elicit chain-of-thought. Minimizing $\mathcal{L}_{\texttt{L}}(\theta)$ via gradient descent adjusts $\theta$ so that the student mimics the teacher's rationales. This is the baseline \emph{one-step} standard knowledge distillation.

\subsection{\undofull}
Standard Knowledge Distillation often leads to a distributional mismatch between the teacher and the student, leading to suboptimal learning \citep{agarwal2024onpolicy, adarsh2024siked}. To counter this, we propose an \emph{iterative} distillation strategy \undo\ that repeatedly identifies the student's errors and lets the teacher produce improved rationales that focus on the \emph{learning gaps} to improve the student's performance. The setup is visualized in \autoref{fig:main-fig}. 

We perform $K$ such iterations, maintaining updated students' parameters $\theta^{(k)}$ after each round. Each step is explained in more detail below:

\paragraph{1) Identify Student Learning Gaps \& Incorporate in Teacher Prompt.}
At each iteration $k$, the student $p_{\texttt{sm}}^{\theta^{(k-1)}}$ answers each problem in the entire validation set $\mathcal{V}$. The student's generation for question $\question_j \in \mathcal{V}$ is denoted as $\tilde{\rationale}_j^{(k,\text{student})}$. We compare the final numeric result within $\tilde{\rationale}_j^{(k,\text{student})}$ to $\answer_j$ and obtain an error signal (e.g., if correct, we provide a score of 1 and 0 otherwise). We collect these labeled pairs 
\[
\bigl(\question_j,\;\tilde{\rationale}_j^{(k,\text{student})},\;\text{score}\bigr),\quad \forall\, j \in \{1,\dots,m\}.
\]
This collection from \emph{all} past iterations (to track improvements over time) will be shown to the teacher in its prompt context, so the teacher can observe the student's progress and mistakes.

\paragraph{2) Regenerate from the Teacher with Gap Analysis.}
Next, we consider the training dataset  $\mathcal{D}_{\texttt{LLM}}$ that we use to improve the student. For each $\question_i \in \mathcal{D}_{\texttt{LLM}}$, we retrieve:
\begin{itemize}
    \item The student’s \emph{previous} generation $\tilde{\rationale}_i^{(k-1,\text{student})}$ (if it exists) or $\tilde{\rationale}_i^{(\text{init})}$,
    \item The teacher’s \emph{previous} rationale $\rationale_i^{(k-1,\text{teacher})}$ used in prior iteration,
\end{itemize}
and construct a prompt that highlights the \emph{gap} between the teacher's prior rationale and the student's generation. The teacher $p_L$ then produces a new rationale
\[
\rationale_i^{(k,\text{teacher})} \;\sim\; p_L\!\Bigl(\cdot \;\Big|\; 
\question_i,\;\tilde{\rationale}_i^{(k-1,\text{student})},\;\rationale_i^{(k-1,\text{teacher})},\;\{\tilde{\rationale}_j^{(k,\text{student})},\, \text{score}_j\}_{j\in \mathcal{V}}
\Bigr),
\]
explicitly aiming to fix or improve upon the student's mistakes. We filter out any rationale whose final numeric result does not match $\answer_i$, retaining a curated set 
\[
\mathcal{D}_{\texttt{LLM}}^{(k)} \;=\; \Bigl\{\bigl(\question_i,\;\rationale_i^{(k,\text{teacher})}\bigr) : \question_i \in \mathcal{D}_{\texttt{LLM}}\Bigr\}.
\]

\paragraph{3) Fine-tune the Student.}
We then use $\mathcal{D}_{\texttt{LLM}}^{(k)}$ to update the student parameters by minimizing the negative log-likelihood of the teacher's new rationales:
\begin{equation*}
\label{eq:iterative_nll}
    \begin{split}
        \mathcal{L}_{\texttt{L}}^{(k)}(\theta)
        \;=\;& -\,\mathbb{E}_{(\question_i,\,\rationale_i^{(k,\text{teacher})}) \,\sim\, \mathcal{D}_{\texttt{LLM}}^{(k)}} 
        \Bigg[
        \sum_{t=1}^{M_i^{(k)}} \log\, p_{\texttt{sm}}^\theta\bigl(\rationale_{i,t}^{(k,\text{teacher})}\mid \rationale_{i,<t}^{(k,\text{teacher})},\,\question_i,\,I\bigr)
        \Bigg],
    \end{split}
\end{equation*}
where $M_i^{(k)}$ is the token length of $\rationale_i^{(k,\text{teacher})}$. We obtain the updated parameters:
\[
\theta^{(k)} \;=\; \arg\min_\theta\;\mathcal{L}_{\texttt{L}}^{(k)}(\theta).
\]

\paragraph{4) Convergence Check.}
We repeat Steps~1--3 until no further improvement is observed on the validation set or until we reach a maximum iteration $K$.

Algorithm~\ref{alg:distillation-as-optimization} summarizes the proposed distillation as optimization strategy. 

\begin{algorithm}[h!]
\caption{\undofull}
\label{alg:distillation-as-optimization}
\begin{algorithmic}[1]
\Require
  \textbf{Teacher:} $p_L(\rationale \mid \question)$\;
  \textbf{Student:} $p_{\texttt{sm}}^{\theta}$\;
  \textbf{Train set:} $\mathcal{D} = \{(\question_i,\answer_i)\}_{i=1}^{n}$\;
  \textbf{Validation set:} $\mathcal{V} = \{(\question_j,\answer_j)\}_{j=1}^{m}$\;
  \textbf{Max Iterations:} $K$\;
  \textbf{Instruction:} $I$ (prompt to elicit CoT).
\State \textbf{Initialize} $\theta^{(0)}$ (e.g., from one-step distillation or a pretrained checkpoint).

\For{$k \gets 1$ to $K$}
  \State \textbf{(1) Identify Student Errors \& Incorporate in Teacher Prompt}
  \State Evaluate $p_{\texttt{sm}}^{\theta^{(k-1)}}$ on each $(\question_j, \answer_j)$ in $\mathcal{V}$:
    \begin{itemize}
      \item Generate student rationale $\tilde{\rationale}_j^{(k,\text{student})}$. 
      \item Extract final answer, compare with $\answer_j$, and record \textit{score}$_j$ (e.g., $1$ if correct, $0$ if wrong).
    \end{itemize}
  \State Collect $\bigl\{\bigl(\question_j,\, \tilde{\rationale}_j^{(k,\text{student})},\, \textit{score}_j\bigr)\bigr\}_{j \in \{1,\dots,m\}}$ for teacher prompts.

  \Statex

  \State \textbf{(2) Regenerate from Teacher with Gap Analysis}
  \For{each $(\question_i, \answer_i) \in \mathcal{D}$}
    \State Retrieve $\tilde{\rationale}_i^{(k-1,\text{student})}$ or $\tilde{\rationale}_i^{(\text{init})}$, 
           and $\rationale_i^{(k-1,\text{teacher})}$ (if they exist).
    \State Construct teacher prompt emphasizing \emph{gaps}:
    \Statex \hspace{\algorithmicindent}
    \(\quad\texttt{prompt} \leftarrow \bigl(\question_i,\; \tilde{\rationale}_i^{(k-1,\text{student})},\;
           \rationale_i^{(k-1,\text{teacher})},\; \{\tilde{\rationale}_j^{(k,\text{student})},\, \textit{score}_j\}_{j \in \mathcal{V}} \bigr).\)
    \State Generate $\rationale_i^{(k,\text{teacher})} \sim p_L(\cdot \mid \texttt{prompt})$.
    \State If final answer in $\rationale_i^{(k,\text{teacher})} \neq \answer_i$, \textbf{discard} the sample.
  \EndFor
  \State Construct $\mathcal{D}_{\texttt{LLM}}^{(k)} = \{(\question_i,\rationale_i^{(k,\text{teacher})}) : \question_i \in \mathcal{D}\text{ (filtered)}\}$.

  \Statex

  \State \textbf{(3) Fine-tune the Student}
  \State Update student parameters:
    \[
      \theta^{(k)} 
      \;\gets\;
      \arg\min_\theta\,
      -\,\mathbb{E}_{(\question_i,\rationale_i^{(k,\text{teacher})}) \,\in\, \mathcal{D}_{\texttt{LLM}}^{(k)}}
      \Biggl[
        \sum_{t=1}^{M_i^{(k)}} \log\, p_{\texttt{sm}}^\theta\bigl(\rationale_{i,t}^{(k,\text{teacher})}
        \mid \rationale_{i,<t}^{(k,\text{teacher})}, \question_i, I\bigr)
      \Biggr].
    \]

  \Statex

  \State \textbf{(4) Convergence Check}
  \State Evaluate $p_{\texttt{sm}}^{\theta^{(k)}}$ on $\mathcal{V}$. If performance (e.g., accuracy) saturates or $k = K$, \textbf{stop}.
\EndFor

\end{algorithmic}
\end{algorithm}

\paragraph{KL Divergence Perspective.}
Each iteration can be viewed as approximating the minimization of the Kullback--Leibler (KL) divergence between the teacher's distribution $p_L(\rationale\mid \question, \text{gap info})$ and the student's distribution $p_{\texttt{sm}}^\theta(\rationale\mid \question)$. If we denote the teacher’s new rationale for question $\question_i$ by $\rationale_i^{(k,\text{teacher})}$, sampled according to $p_L(\rationale \mid \question_i, \Delta_i^{(k)})$ (where $\Delta_i^{(k)}$ encapsulates the gap information, e.g., old student and teacher rationales, validation errors, etc.), then the KL divergence is
\[
\mathrm{KL}\Bigl(p_L(\rationale\mid \question_i, \Delta_i^{(k)}) \;\|\; p_{\texttt{sm}}^\theta(\rationale\mid \question_i)\Bigr)
\;=\; 
\mathbb{E}_{\rationale \sim p_L(\cdot\mid \question_i,\Delta_i^{(k)})}
\Bigl[\log\,p_L(\rationale\mid \question_i,\Delta_i^{(k)}) 
\;-\; \log\,p_{\texttt{sm}}^\theta(\rationale\mid \question_i)\Bigr].
\]
Since the term $\log\,p_L(\rationale\mid \question_i,\Delta_i^{(k)})$ is independent of $\theta$, minimizing the KL is equivalent (up to constants) to maximizing
\(\mathbb{E}_{\rationale \sim p_L}[\log p_{\texttt{sm}}^\theta(\rationale\mid \question_i)],\)
which matches the negative log-likelihood objective used in Eq.~\eqref{eq:iterative_nll} when we replace the teacher's distribution by its empirical samples $\rationale_i^{(k,\text{teacher})}$. Iteration-by-iteration, the teacher highlights gaps by conditioning on the student's errors (\emph{via} $\Delta_i^{(k)}$), so the newly generated rationales better target exactly where $p_{\texttt{sm}}^\theta$ remains deficient. Consequently, with repeated minimization of this KL divergence for ``hard'' or incorrectly answered questions, the student's distribution $p_{\texttt{sm}}^\theta$ is successively refined, thereby improving overall performance.

\section{Experimental Details}
\subsection{Dataset}

We utilized Numina Math dataset \citep{numina_math_datasets} to fine-tune the student models. We filtered out only math-word-problem "question-type" and further did some preprocessing to remove smaples where the answer was "proof", "notfound", empty or the problem/solution was invalid. Since it does not include a validation set, we partitioned the training set to create a held-out validation set. We used 20 samples sampled randomly from the training set to create a validation set. Note that since validation set is always kept in the context of the teacher model and grows linearly with the iterations, we kept the validation set to fit in the context window even after $K$ iterations. For the test set, we aggregated test sets from various other mathematical datasets, including GSM8K \citep{cobbe2021gsm8k}, MMLU Pro \citep{wang2024mmlupro}, MATH 500 \citep{lightman2024lets}, and SVAMP \citep{Patel2021AreNM}. 
The dataset distribution is provided in
\autoref{tab:dataset-distribution-table}.
Finally, we also test our approach on two out-of-domain commonsense reasoning datasets of StrategyQA \cite{strategyQA} (687 test samples \footnote{Test samples taken from this repo: \url{https://huggingface.co/datasets/ChilleD/StrategyQA}}) and TheoremQA \cite{chen-etal-2023-theoremqa} (800 test samples).  We use 4-shot CoT for StrategyQA and 5-shot CoT for TheoremQA evaluation.

\begin{table}[ht!]
\centering
\begin{tabular}{lrrrr}
\toprule
\multirow{2}{*}{\textbf{Dataset}} 
  & \multicolumn{2}{c}{\textbf{Training Set (N = 550,000)}} 
  & \multicolumn{2}{c}{\textbf{Test Set (N = 4170)}} \\
\cmidrule(lr){2-3} \cmidrule(lr){4-5}
 & \textbf{\# Samples} & \textbf{Dist. (\%)} & \textbf{\# Samples} & \textbf{Dist. (\%)} \\
\midrule
\textbf{NuminaMath} \citep{numina_math_datasets} & 550,000 & 100\%  & -  & - \\
\textbf{Math500} \citep{lightman2024lets}   & - & -  & 500  & 12\% \\
\textbf{GSM8K}   \citep{cobbe2021gsm8k}   & - & - & 1319 &  32\%\\
\textbf{MMLU PRO} \citep{wang2024mmlupro}   & - & - & 1351 & 32\% \\
\textbf{SVAMP}  \citep{Patel2021AreNM}   & - & - & 1000 & 24\% \\
\bottomrule
\end{tabular}
\vspace{1\baselineskip}
\caption{The distribution of datasets used in our work.}
\vspace{1\baselineskip}
\label{tab:dataset-distribution-table}
\end{table}
\vspace{1\baselineskip}

\begin{table}[t]
\centering
\vspace{6pt}  % <<-- This adds some vertical space between the caption and the table
\begin{tabular}{llccccc}
\toprule
\textbf{Model} & \textbf{Iter.} & \textbf{GSM8K} & \textbf{MATH} & \textbf{MMLU PRO} & \textbf{SVAMP} & \textbf{Average} \\
\midrule
%---------------- QWEN ----------------
\multirow{3}{*}{\rotatebox[origin=c]{90}{\textbf{Qwen}}}
 & 1 
   & 50.95 
   & 32.80
   & 14.51
   & 76.70
   & 43.14 \\
 & 2 
   & \underline{53.22} \hlg{{\scalebox{0.7}{($\uparrow$+2.27)}}}
   & \underline{35.80} \hlg{{\scalebox{0.7}{($\uparrow$+3.00)}}}
   & 15.40 \hlg{{\scalebox{0.7}{($\uparrow$+0.89)}}}
   & \underline{84.20} \hlg{{\scalebox{0.7}{($\uparrow$+7.50)}}}
   & \underline{46.31} \hlg{{\scalebox{0.7}{($\uparrow$+3.17)}}} \\
 & 3 
   & \underline{55.04} \hlg{{\scalebox{0.7}{($\uparrow$+4.09)}}}
   & \underline{39.00} \hlg{{\scalebox{0.7}{($\uparrow$+6.20)}}}
   & 15.17 \hlg{{\scalebox{0.7}{($\uparrow$+0.66)}}}
   & \underline{87.40} \hlg{{\scalebox{0.7}{($\uparrow$+10.70)}}}
   & \underline{47.96} \hlg{{\scalebox{0.7}{($\uparrow$+4.82)}}} \\
   \\
\midrule
%---------------- SmolLM2 ----------------
\multirow{3}{*}{\rotatebox[origin=c]{90}{\textbf{SmolLM}}}
 & 1 
   & 53.60 
   & 30.60
   & 13.47
   & 85.50
   & 45.49 \\
 & 2 
   & \underline{55.72} \hlg{{\scalebox{0.7}{($\uparrow$+2.12)}}}
   & \underline{34.20} \hlg{{\scalebox{0.7}{($\uparrow$+3.60)}}}
   & \underline{15.03} \hlg{{\scalebox{0.7}{($\uparrow$+1.56)}}}
   & \underline{87.20} \hlg{{\scalebox{0.7}{($\uparrow$+1.70)}}}
   & 47.51 \hlg{{\scalebox{0.7}{($\uparrow$+2.02)}}} \\
 & 3 
   & \underline{56.93} \hlg{{\scalebox{0.7}{($\uparrow$+3.33)}}}
   & \underline{38.00} \hlg{{\scalebox{0.7}{($\uparrow$+7.40)}}}
   & \underline{16.21} \hlg{{\scalebox{0.7}{($\uparrow$+2.74)}}}
   & \underline{87.20} \hlg{{\scalebox{0.7}{($\uparrow$+1.70)}}}
   & \underline{48.70} \hlg{{\scalebox{0.7}{($\uparrow$+3.21)}}} \\
   \\
\midrule
%---------------- LLAMA ----------------
\multirow{3}{*}{\rotatebox[origin=c]{90}{\textbf{Llama}}}
 & 1 
   & 34.72 
   & 20.60
   &  9.74
   & 79.50
   & 35.59 \\
 & 2 
   & \underline{39.34} \hlg{{\scalebox{0.7}{($\uparrow$+4.62)}}}
   & \underline{24.80} \hlg{{\scalebox{0.7}{($\uparrow$+4.20)}}}
   & \underline{12.40} \hlg{{\scalebox{0.7}{($\uparrow$+2.66)}}}
   & 78.50 \hld{{\scalebox{0.7}{($\downarrow$-1.00)}}}
   & \underline{38.27} \hlg{{\scalebox{0.7}{($\uparrow$+2.68)}}} \\
 & 3 
   & \underline{43.96} \hlg{{\scalebox{0.7}{($\uparrow$+9.24)}}}
   & \underline{27.00} \hlg{{\scalebox{0.7}{($\uparrow$+6.40)}}}
   & \underline{15.06} \hlg{{\scalebox{0.7}{($\uparrow$+5.32)}}}
   & 80.50 \hlg{{\scalebox{0.7}{($\uparrow$+1.00)}}}
   & \underline{40.95} \hlg{{\scalebox{0.7}{($\uparrow$+5.36)}}} \\
\bottomrule
\end{tabular}
\vspace{1\baselineskip}
\caption{Accuracy (\texttt{maj@1}) comparison for three student models (Qwen2.5 1.5B, SmolLM2 1.7B, and Llama-3.2 1B) over different iterations (Iter.). 
$\uparrow$ indicates the gain (or loss, $\downarrow$) relative to Iteration1  of the same model, which is the baseline standard distillation. \underline{Underline} indicates statistical significance with $p < 0.05$ compared to the baseline. }
\vspace{1\baselineskip}
\label{tab:main-table}
\end{table}
\vspace{1\baselineskip}

\subsection{Models}
We employed \texttt{llama-3.2-70B} \citep{llama3} as the teacher model and selected three student models—\texttt{llama-3.3-1B}, \texttt{qwen-2.5-1.5B} \citep{qwen}, and \texttt{SmolLM2-1.7B} \citep{allal2025smollm2}. The student models were chosen for their diversity in size and performance. Across all experiments, the number of iterations ($K$) reached saturation at 4, meaning  the optimal results were observed at $K=3$. For each student model training run, we utilized 5 epochs for the initial iteration ($K=1$) and reduced it to 3 epochs for subsequent iterations. Training each student model required approximately 10 GPU hours per epoch on H100 GPUs. Inference runs on the training data for the student models consumed approximately 240 GPU hours on GH200 GPUs, whereas generating data with the teacher model took approximately 2,800 GPU hours (GH200s) per iteration. The temperature was set at $0.7$ and inference was done using the VLLM library \citep{vllm}.

\section{Results and Discussion}

\paragraph{Iterative distillation improves student performance.}
Table~\ref{tab:main-table} presents the results of three different student models (\textbf{Qwen2.5 1.5B}, \textbf{SmolLM2 1.7B}, and \textbf{Llama-3.2 1B}) over three iterations (K) of our iterative distillation procedure \undo. We report performance on four mathematical reasoning datasets: \texttt{GSM8K}, \texttt{MATH}, \texttt{MMLU PRO}, and \texttt{SVAMP}, along with an \texttt{Average} score across these tasks. We observe a consistent improvement over all datasets across all models. For instance, under \textbf{Qwen2.5 1.5B}, the performance on GSM8K increases from $50.95$ at iteration~1 to $55.04$ at iteration~3. A similar trend is observed for the MATH dataset, which improves from $32.80$ $\rightarrow$ $39.00$, and on SVAMP which increases from $76.70$ $\rightarrow$ $87.40$. SmolLM2 and Llama-3.2 also follow the same upward trajectory. Notably, \textbf{SmolLM2} gains $+7.40$ points on MATH by iteration~3 (from $30.60$ $\rightarrow$ $38.00$) and $+2.74$ points on MMLU PRO (from $13.47$ $\rightarrow$ $16.21$). \textbf{Llama-3.2} sees gains of $+9.24$ and $+8.40$ points on GSM8K and MATH, respectively, moving from $34.72$ $\rightarrow$ $43.96$ on GSM8K and $20.60$ $\rightarrow$ $29.00$ on MATH. Overall, each model consistently benefits from iterative distillation, achieving higher accuracy on each dataset from iteration~1 to iteration~3.

\begin{figure}[t]
    \centering
    \includegraphics[width=0.99\linewidth]{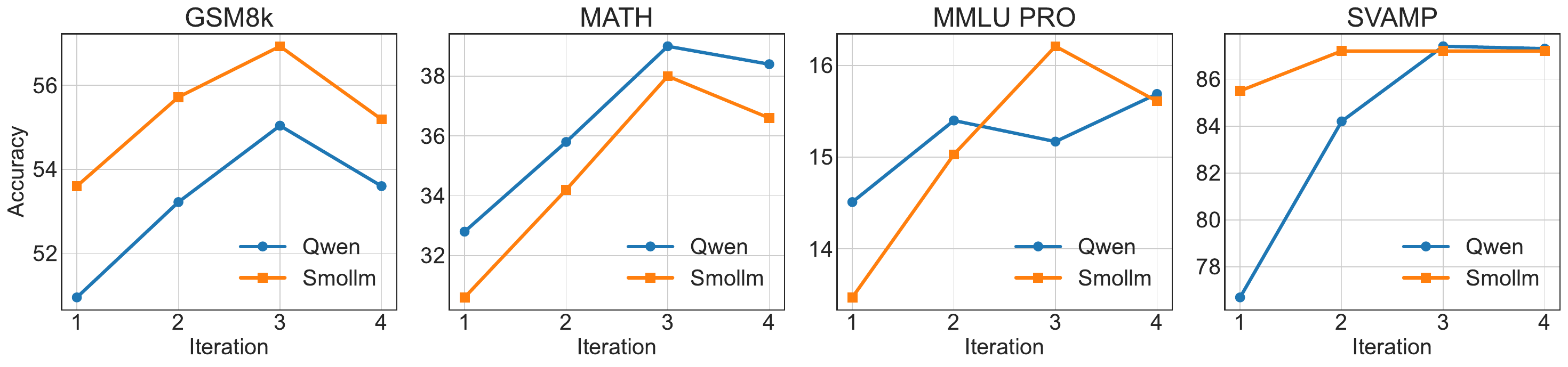}
    \vspace{1\baselineskip}
    \caption{Accuracy (maj@1) for Qwen and SmolLM2 models on GSM8K, MATH, MMLU Pro and SVAMP datasets for 4 iterations.}
    \label{fig:stopping-criteria}
\end{figure}

\paragraph{Deciding when to stop}
Figure~\ref{fig:stopping-criteria} visualizes the performance of Qwen2.5 and SmolLM2 across four iterations on the same four benchmarks. From this plot, we see a strong upward trend through the first three iterations, aligning with the numerical gains reported in Table~\ref{tab:main-table}. However, at iteration~4, there is a slight decrease in accuracy for both models: Qwen2.5 drops from $55.04$ $\rightarrow$ $53.60$ on GSM8K, and SmolLM2 dips from $56.93$ $\rightarrow$ $55.19$ on the same dataset. Since these iteration~4 numbers are lower than iteration~3 results, we conclude that the model has converged and additional regeneration from the teacher is no longer beneficial, prompting us to stop further distillation steps at iteration~3.

\vspace{1\baselineskip}

\paragraph{Can standard distillation trained for same number of epochs perform similar to \undo? }

\begin{figure}
    \centering
    \includegraphics[width=0.99\linewidth]{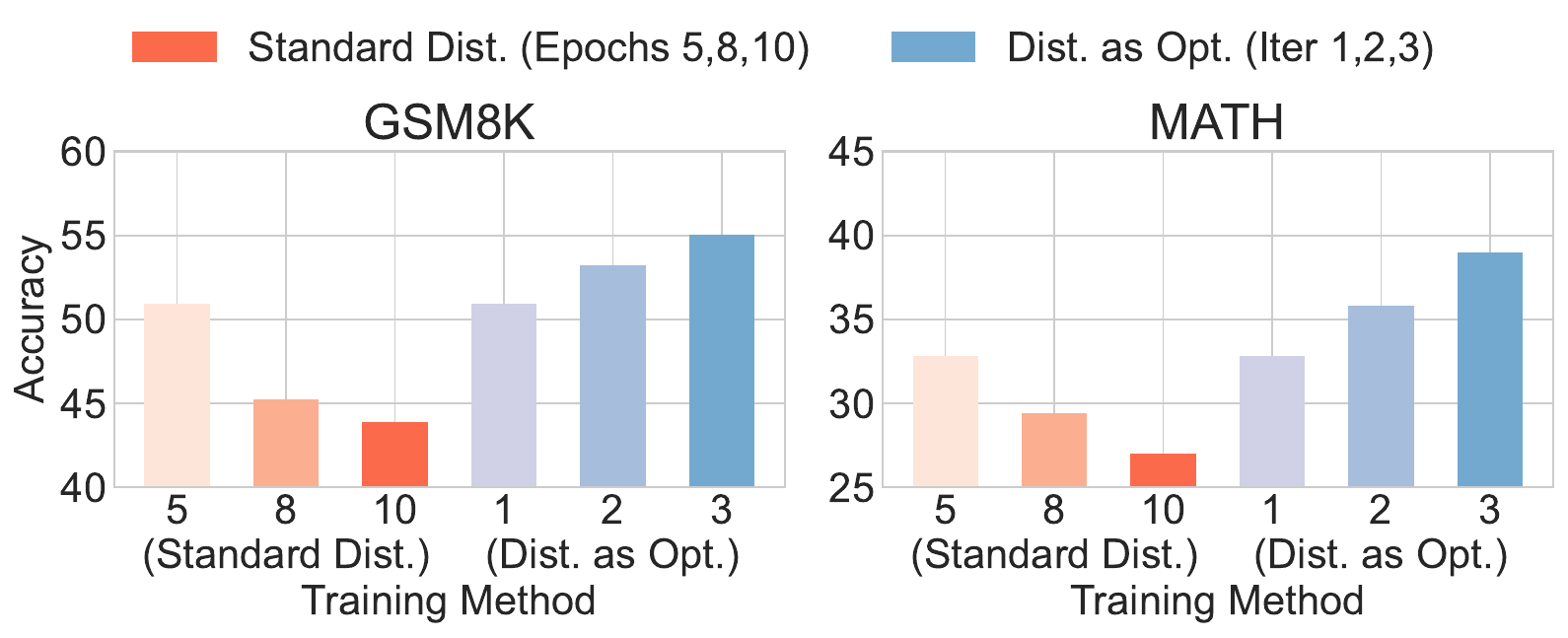}
    \vspace{1\baselineskip}
    \caption{Accuracy comparison of training standard distillation and \undo\ for the equal number of epochs. Note that iteration 1 for \undo\ is same as epoch 5 for standard distillation.}
    \label{fig:same-train-epoch}
\end{figure}
\vspace{1\baselineskip}

Figure~\ref{fig:same-train-epoch} illustrates that simply continuing to train Qwen (\emph{standard distillation} approach) for more epochs beyond \texttt{Iteration 1} results in diminished returns and overfitting. Concretely, the model trained for 10 epochs scores only \(43.90\%\) on GSM8k, \(27.00\%\) on MATH, \(13.82\%\) on MMLU PRO, and \(72.90\%\) on SVAMP, for an average of \(39.41\%\)---all lower than the baseline \texttt{iteration 1} model. In contrast, our proposed approach \undo\ improves performance in each iteration: on GSM8K, for instance, the student goes from \(50.95\%\) at Iter1 to \(55.04\%\) at Iter3; on MATH from \(32.80\%\) to \(39.00\%\); on MMLU PRO from \(14.51\%\) to \(15.17\%\); on SVAMP from \(76.70\%\) to \(87.40\%\); culminating in an average gain from \(43.14\%\) to \(47.96\%\), all trained for the same number of epochs. These results confirm that simply increasing training epochs risks overfitting and declining performance, while iterative distillation actively pushes the student model toward the teacher’s  reasoning distribution. 

\begin{table}[t]
\centering
\vspace{6pt}  % <<-- This adds some vertical space between the caption and the table
\begin{tabular}{llccccc}
\toprule
\textbf{} & \textbf{Teacher} & \textbf{GSM8K} & \textbf{MATH} & \textbf{MMLU PRO} & \textbf{SVAMP} & \textbf{Average} \\
\midrule
%---------------- QWEN ----------------
\multirow{3}{*}{\rotatebox[origin=c]{90}{\textbf{Qwen}}}
 & Baseline 
   & 50.95 
   & 32.80
   & 14.51
   & 76.70
   & 43.14 \\
 & Qwen 
   & \underline{56.10} \hlg{{\scalebox{0.7}{($\uparrow$+5.15)}}}
   & \underline{39.40} \hlg{{\scalebox{0.7}{($\uparrow$+6.60)}}}
   & 15.98 \hlg{{\scalebox{0.7}{($\uparrow$+1.47)}}}
   & \underline{86.50} \hlg{{\scalebox{0.7}{($\uparrow$+9.80)}}}
   & \underline{48.39} \hlg{{\scalebox{0.7}{($\uparrow$+5.52)}}} \\
 & SmolLM2 
   & \underline{54.00} \hlg{{\scalebox{0.7}{($\uparrow$+3.05)}}}
   & 34.00 \hlg{{\scalebox{0.7}{($\uparrow$+1.20)}}}
   & 14.35 \hld{{\scalebox{0.7}{($\downarrow$-0.16)}}}
   & \underline{86.50} \hlg{{\scalebox{0.7}{($\uparrow$+9.80)}}}
   & \underline{46.57} \hlg{{\scalebox{0.7}{($\uparrow$+3.43)}}} \\
\midrule
%---------------- SmolLM2 ----------------
\multirow{3}{*}{\rotatebox[origin=c]{90}{\textbf{SmolLM}}}
 & Baseline 
   & 53.60 
   & 30.60
   & 13.47
   & 85.50
   & 45.49 \\
 & Qwen 
   & \underline{56.86} \hlg{{\scalebox{0.7}{($\uparrow$+3.26)}}}
   & \underline{32.80} \hlg{{\scalebox{0.7}{($\uparrow$+2.20)}}}
   & 14.95 \hlg{{\scalebox{0.7}{($\uparrow$+1.48)}}}
   & 85.60 \hlg{{\scalebox{0.7}{($\uparrow$+0.10)}}}
   & 47.29 \hlg{{\scalebox{0.7}{($\uparrow$+1.80)}}} \\
 & SmolLM2 
   & \underline{56.86} \hlg{{\scalebox{0.7}{($\uparrow$+3.26)}}}
   & \underline{33.60} \hlg{{\scalebox{0.7}{($\uparrow$+3.00)}}}
   & \underline{15.00} \hlg{{\scalebox{0.7}{($\uparrow$+1.53)}}}
   & 86.80 \hlg{{\scalebox{0.7}{($\uparrow$+1.30)}}}
   & \underline{47.72} \hlg{{\scalebox{0.7}{($\uparrow$+2.23)}}} \\
   \\
\bottomrule
\end{tabular}
\vspace{1\baselineskip}
\caption{Comparison of \texttt{maj@1} accuracy for Qwen and SmolLM across four benchmarks (GSM8K, MATH, MMLU PRO, SVAMP) and the resulting average score. Each model can be trained on data generated by its own teacher or a cross-teacher (e.g., Qwen data for SmolLM2). Gains (\(\uparrow\) or \(\downarrow\)) are shown relative to the baseline (standard distillation). \underline{Underline} indicates statistical significance with $p < 0.05$ compared to the baseline.}
\vspace{1\baselineskip}
\label{tab:ablation-table}
\end{table}
\vspace{1\baselineskip}

\paragraph{Does \undo\ create a better teacher?}
Table~\ref{tab:ablation-table} shows that once we complete our iterative distillation procedure and use the final teacher generated data to train student models as standard distillation for 5 epochs, we achieve substantial gains over the baseline. Specifically, Qwen trained on the final iteration teacher generations from a Qwen-Llama 70B interaction, improves by more than +5 points on GSM8K (from 50.95 $\rightarrow$ 56.10) and +9.80 points on SVAMP (from 76.70 $\rightarrow$ 86.50), culminating in an overall average increase of +5.52. Even when Qwen is trained on teacher generations of a SmolLM2-Llama 70B interaction, it still outperforms the baseline (e.g., +3.43 on average), though slightly less than with Qwen’s own teacher data. This suggests that the teacher model—refined through \undo—produces data that is more \emph{personalized} to the student’s weaknesses, thus effectively boosting final performance. We also observe similar trends for the SmolLM2 model, which gains up to +3.26 accuracy on GSM8K (from 53.60 $\rightarrow$ 56.86) and +1.53 on MMLU PRO (13.47 $\rightarrow$ 15.00) when trained on the teacher data from a SmolLM2-Llama 70B interaction. Similarly, SmolLM2 also benefits from teacher data of a Qwen-Llama 70B interaction gaining almost +2 points on average over the baseline. The improvements indicate that the teacher–student pair evolves into a tightly coupled relationship, outperforming a standard distillation baseline where the teacher lacks any prior knowledge of the student’s mistakes.

\paragraph{Is training with \undo\ make the student overfit to the task?}
We test if the student is overfitting when trained with \undo\ compared to the standard distillation by testing on two out of domain datasets: StrategyQA \citep{strategyQA} and TheoremQA \citep{chen-etal-2023-theoremqa}. 
Table~\ref{tab:two_datasets} shows that both Qwen and Smolm consistently improve their performance from Iteration 1 to Iteration 3 on the \textbf{StrategyQA} and \textbf{Theorem QA} benchmarks. For instance, Qwen increases by 2.50\% on StrategyQA and 0.625\% on Theorem QA, while Smolm gains 2.80\% and 0.63\% respectively. These consistent improvements across both datasets indicate that the models are not overfitted to the mathematical reasoning tasks. Rather, they have acquired better reasoning skills which are also transferrable to out-of-domain datasets.

\begin{table}[ht]
\centering
\begin{tabular}{lcc|cc}
\toprule
 & \multicolumn{2}{c}{\textbf{StrategyQA}} & \multicolumn{2}{c}{\textbf{Theorem QA}} \\
\cmidrule(lr){2-3} \cmidrule(lr){4-5}
\textbf{Model} & \textbf{Standard Dist.} & \textbf{\undo} & \textbf{Standard Dist.} & \textbf{\undo} \\
\midrule
\textbf{Qwen} & 12.50\% & 15.00\%\ \hlg{{\scriptsize($\uparrow$+2.50)}} & 7.5\% & 8.125\%\ \hlg{{\scriptsize($\uparrow$+0.625)}} \\
\textbf{SmolLM2} & 10.20\% & 13.00\%\ \hlg{{\scriptsize($\uparrow$+2.80)}} & 5.62\% & 6.25\%\ \hlg{{\scriptsize($\uparrow$+0.63)}} \\
\bottomrule
\end{tabular}
\vspace{1\baselineskip}
\caption{Performance (\%) on \textbf{StrategyQA} and \textbf{Theorem QA} benchmarks for Qwen and SmolLM2 models. Standard Distillation represents the results from Iteration 1 while \undo\ represents the results from Iteration 3 from \autoref{tab:main-table}.
Gains ($\uparrow$) are shown in parentheses.}
\vspace{1\baselineskip}
\label{tab:two_datasets}
\end{table}
\vspace{1\baselineskip}

\section{Conclusion}

We introduced an iterative distillation paradigm \undo\ aimed at bridging the performance gap between large teacher models and smaller student models more effectively than conventional one-shot distillation. By repeatedly identifying the student’s errors, regenerating teacher rationales that address these \emph{learning gaps}, and re-training the student on this tailored data, our method yields consistent improvements on challenging mathematical benchmarks. Experimental results indicate that \undo\ not only avoids overfitting often seen in prolonged standard distillation, but also makes the teacher more attuned to student-specific weaknesses, effectively “personalizing” the teacher’s rationales.

Empirically, we observed that final teacher-generated data—after multiple rounds of iteration—significantly enhances student performance, with gains of upto 20\% compared to the baseline. Even cross-teacher data (e.g., teacher data for one student used on another) shows improved results over standard approaches. These findings underscore that incorporating iterative feedback loops between the teacher and the student leads to a more specialized and effective transfer of reasoning knowledge.

\bibliography{colm2025_conference}
\bibliographystyle{colm2025_conference}

\appendix
\section{Appendix}

% \begin{listing*}
\begin{small}
\vspace{2mm}
\label{fig:prompt-iter1}
\begin{center}
\textbf{Prompt file given to the teacher in the first iteration.}
\end{center}
\captionsetup{justification=centering, labelfont=bf, font=small}
\begin{minted}[fontsize=\footnotesize, frame=lines, framesep=2mm, baselinestretch=1.2, breaklines, breaksymbolleft={}, breaksymbolright={}]{text}
You are an AI tutor tasked with improving a student's understanding of mathematical problem-solving. You will be given a question, a teacher's answer, a student's answer, and a score. Your job is to analyze these inputs and create a new answer that will help the student learn better.

Here are some examples of the task:

### question:
Jim spends 8 hours scuba diving. In that time he finds a treasure chest with 100 gold coins in it. He also finds some smaller bags that have half as much gold each. He finds 25 gold coins per hour. How many smaller bags did he find?
### teacher answer:
Jim finds 25 gold coins per hour, and he spends 8 hours scuba diving, so he finds a total of 25 * 8 = 200 gold coins. 
He finds a treasure chest with 100 gold coins in it, so the remaining gold coins he finds are 200 - 100 = 100 gold coins. 
Since the smaller bags have half as much gold as the treasure chest, each smaller bag has 100 / 2 = 50 gold coins. 
Therefore, the number of smaller bags he finds is 100 / 50 = $\boxed{2}$ smaller bags.
Final Answer: 2
### student answer:
To determine how many smaller bags Jim found, we need to follow these steps: 
1. **Calculate the total amount of gold coins Jim finds over the 8 hours:** Jim finds 25 gold coins per hour for 8 hours. \[ 25 \text{ coins/hour} \times 8 \text{ hours} = 200 \text{ coins} \] 
2. **Determine the amount of gold coins in the smaller bags:** Let \( x \) be the number of smaller bags. Each smaller bag has half as much gold as a treasure chest. Since a treasure chest has 100 gold coins, each smaller bag has: \[ \frac{100 \text{ coins}}{2} = 50 \text{ coins} \] Therefore, the total amount of gold coins in the smaller bags is: \[ 50x \text{ coins} \] 
3. **Set up the equation for the total gold coins found:** The total gold coins found by Jim is the sum of the gold coins in the treasure chest and the gold coins in the smaller bags. According to the problem, this total is 200 coins. \[ 100 \text{ coins} + 50x \text{ coins} = 200 \text{ coins} \] 
4. **Solve for \( x \):** Subtract 100 coins from both sides of the equation: \[ 50x = 100 \] Divide both sides by 50: \[ x = \frac{100}{50} = 2 \] Thus, Jim found \(\boxed{2}\) smaller bags.
Final Answer: 2
### score:
1

------------similar such examples from the validation set------------


First, carefully analyze the student's answer. Compare it to the teacher's answer and identify any mistakes or areas where the student's reasoning could be improved. Consider the following:

1. Did the student understand the problem correctly?
2. Did they use the right approach to solve the problem?
3. Are there any calculation errors?
4. Is their reasoning clear and logical?
5. Did they miss any important steps?

Next, craft a new answer that addresses the student's misunderstandings or reinforces correct thinking. Your new answer should:

1. Use clear, step-by-step reasoning
2. Explain any concepts the student may have misunderstood
3. Provide additional context or examples if necessary
6. Use the same calculation format as the teacher's answer. If the teacher answer involves latex and involves terms like \frac, \pi make sure to extract the answer with the necessary latex keywords. 
5. All the teacher's final answers end inside "\boxed{}". Ensure, that your answers also follow this format.
6. Lead to the correct final answer

Write your new answer using the following format:

### new_answer
[Step-by-step reasoning with calculations in the format shown above]
Final Answer: [Correct numerical/latex answer]


Remember, your goal is to help the student learn and improve their problem-solving skills. Focus on explaining the reasoning clearly and addressing any specific issues in the student's original answer.
\end{minted}
\end{small}
% \caption{Prompt file given to the teacher in the first iteration.}
% \end{listing*}

% \begin{figure}[H]
\begin{small}
\vspace{2mm}
\begin{center}
\textbf{Prompt file given to the teacher in the second iteration.}
\end{center}
\captionsetup{justification=centering, labelfont=bf, font=small}
\begin{minted}[fontsize=\footnotesize, frame=lines, framesep=2mm, baselinestretch=1.2, breaklines, breaksymbolleft={}, breaksymbolright={}]{text}
You are an AI tutor tasked with improving a student's understanding of mathematical problem-solving. You will be given a question, a teacher's answer, a student's answer, and a score. Your job is to analyze these inputs and create a new answer that will help the student learn better.

Here are some examples of the task:

### question:
Jim spends 8 hours scuba diving. In that time he finds a treasure chest with 100 gold coins in it. He also finds some smaller bags that have half as much gold each. He finds 25 gold coins per hour. How many smaller bags did he find?
### ITERATION 1:
### teacher answer:
Jim finds 25 gold coins per hour, and he spends 8 hours scuba diving, so he finds a total of 25 * 8 = 200 gold coins. 
He finds a treasure chest with 100 gold coins in it, so the remaining gold coins he finds are 200 - 100 = 100 gold coins. 
Since the smaller bags have half as much gold as the treasure chest, each smaller bag has 100 / 2 = 50 gold coins. 
Therefore, the number of smaller bags he finds is 100 / 50 = $\boxed{2}$ smaller bags.
Final Answer: 2
### student answer:
To determine how many smaller bags Jim found, we need to follow these steps: 
1. **Calculate the total amount of gold coins Jim finds over the 8 hours:** Jim finds 25 gold coins per hour for 8 hours. \[ 25 \text{ coins/hour} \times 8 \text{ hours} = 200 \text{ coins} \] 
2. **Determine the amount of gold coins in the smaller bags:** Let \( x \) be the number of smaller bags. Each smaller bag has half as much gold as a treasure chest. Since a treasure chest has 100 gold coins, each smaller bag has: \[ \frac{100 \text{ coins}}{2} = 50 \text{ coins} \] Therefore, the total amount of gold coins in the smaller bags is: \[ 50x \text{ coins} \] 
3. **Set up the equation for the total gold coins found:** The total gold coins found by Jim is the sum of the gold coins in the treasure chest and the gold coins in the smaller bags. According to the problem, this total is 200 coins. \[ 100 \text{ coins} + 50x \text{ coins} = 200 \text{ coins} \] 
4. **Solve for \( x \):** Subtract 100 coins from both sides of the equation: \[ 50x = 100 \] Divide both sides by 50: \[ x = \frac{100}{50} = 2 \] Thus, Jim found \(\boxed{2}\) smaller bags.
Final Answer: 2
### score:
1
### ITERATION 2:
### teacher answer:
To determine how many smaller bags Jim found, we need to follow these steps: 
1. **Calculate the total amount of gold coins Jim finds over 8 hours:** Jim finds 25 gold coins per hour for 8 hours. The total gold coins he finds is: \[ 25 \times 8 = 200 \text{ gold coins} \] 
2. **Determine the amount of gold coins in the smaller bags:** The problem states that each smaller bag has half as much gold as a treasure chest. Since a treasure chest contains 100 gold coins, a smaller bag will have: \[ \frac{100}{2} = 50 \text{ gold coins} \] 
3. **Set up an equation to find the number of smaller bags \( x \):** The total amount of gold coins found in the smaller bags over 8 hours should equal the total gold coins found minus the gold coins in the treasure chest. Let \( x \) be the number of smaller bags found. The total gold coins found in the smaller bags is: \[ 50x \] This total must equal the remaining gold coins after subtracting the treasure chest: \[ 50x = 200 - 100 \] \[ 50x = 100 \] 
4. **Solve for \( x \):** To find \( x \), divide both sides of the equation by 50: \[ x = \frac{100}{50} = 2 \] Thus, the correct answer is: \boxed{2} Final Answer: \boxed{2}
Final Answer: 2
### student answer:
To find the number of smaller bags Jim found, we first calculate the total amount of gold coins he has. 
Since he finds 25 gold coins per hour for 8 hours, the total amount of gold coins he finds is $25 \times 8 = 200$ gold coins. Next, we determine the amount of gold coins in each smaller bag. Each smaller bag has half as much gold as a treasure chest, which is $\frac{100}{2} = 50$ gold coins. 
Now, we divide the total amount of gold coins found by the amount in each smaller bag to find the number of smaller bags: $\frac{200}{50} = 4$. Therefore, Jim found $\boxed{4}$ smaller bags. Final Answer: \boxed{4}.
Final Answer: 4
### score:
0


------------similar such examples from the validation set------------

First, carefully analyze the student's answer. Compare it to the teacher's answer and identify any mistakes or areas where the student's reasoning could be improved. Consider the following:

1. Did the student understand the problem correctly?
2. Did they use the right approach to solve the problem?
3. Are there any calculation errors?
4. Is their reasoning clear and logical?
5. Did they miss any important steps?

Next, craft a new answer that addresses the student's misunderstandings or reinforces correct thinking. Your new answer should:

1. Use clear, step-by-step reasoning
2. Explain any concepts the student may have misunderstood
3. Provide additional context or examples if necessary
6. Use the same calculation format as the teacher's answer. If the teacher answer involves latex and involves terms like \frac, \pi make sure to extract the answer with the necessary latex keywords. 
5. All the teacher's final answers end inside "\boxed{}". Ensure, that your answers also follow this format.
6. Lead to the correct final answer

Write your new answer using the following format:

### new_answer
[Step-by-step reasoning with calculations in the format shown above]
Final Answer: [Correct numerical/latex answer]


Remember, your goal is to help the student learn and improve their problem-solving skills. Focus on explaining the reasoning clearly and addressing any specific issues in the student's original answer.
\end{minted}
\label{fig:prompt-iter2}
\end{small}
% \caption{Prompt file given to the teacher in the second iteration.}
\end{document}